\documentclass[11pt,a4paper]{article}
\usepackage{times,latexsym}
\usepackage{url}
\usepackage[T1]{fontenc}

\usepackage[acceptedWithA]{tacl2021v1}

\usepackage{soul}
\usepackage{amsmath, amsfonts}
\usepackage[capitalise,noabbrev]{cleveref}
\usepackage{paralist}
\usepackage{enumitem}
\usepackage[whole]{bxcjkjatype}

\usepackage{booktabs}
\usepackage{tabularx}
\usepackage{array} 
\usepackage{multirow}
\usepackage{calc}
\usepackage{makecell}

\usepackage{placeins} 

\usepackage{algorithm,algpseudocode}
\algnewcommand\Input{\item[\textbf{Input:}]}
\algnewcommand\Output{\item[\textbf{Output:}]}
\algrenewcommand\algorithmicindent{1.0em}
\algnewcommand\Var[1]{\textit{#1}}
\algnewcommand\BVar[1]{\textbf{\textit{#1}}}

\usepackage{xspace,mfirstuc,tabulary}

\newif\iftaclinstructions
\taclinstructionsfalse 
\iftaclinstructions
\renewcommand{\confidential}{}
\renewcommand{\anonsubtext}{(No author info supplied here, for consistency with
TACL-submission anonymization requirements)}
\newcommand{\instr}
\fi

\iftaclpubformat 

\else

\fi

\title{Towards Automated Lexicography:\\Generating and Evaluating Definitions for Learner's Dictionaries}

\author{
  \textbf{Yusuke Ide${}^{1,2}$ \; Adam Nohejl${}^{2,1}$ \; Joshua Tanner${}^{3}$} \\
  \textbf{Hitomi Yanaka${}^{2,4,5}$ \; Christopher Lindsay${}^{6}$ \; Taro Watanabe${}^{1}$ \vspace{4pt}} \\
  $^1$Nara Institute of Science and Technology \; $^2$RIKEN \; $^3$Resolve Research \\
  $^4$The University of Tokyo \; $^5$Tohoku University \; $^6$Serpenti Sei Japan \vspace{3pt} \\
  \fontsize{10.5pt}{10.5pt}\selectfont \texttt{ide.yusuke.ja6@is.naist.jp, adam.nohejl@riken.jp, josh@rslv.dev} \\
  \fontsize{10.5pt}{10.5pt}\selectfont \texttt{hyanaka@is.s.u-tokyo.ac.jp, chris@serpentisei.com, taro@is.naist.jp}
}

\date{}

\begin{document}

\newlength{\CJKw}
\newcommand{\kernCJK}{\settowidth{\CJKw}{一}\nobreak\kern -0.5\CJKw}

\newlength{\minusw}
\newcommand{\niceminus}{\textminus\kern 0.05em}
\newcommand{\nominus}{\phantom{\niceminus}}
\newcommand{\tableminus}{\settowidth{\minusw}{\niceminus}{\kern -\minusw}\niceminus} 


\maketitle
\begin{abstract}
    Dictionary definitions are an essential resource for learning word senses, but manually creating them is costly.
    We thus study dictionary definition generation (DDG), i.e., the generation of non-contextualized definitions for given headwords.
    Specifically, we address learner's dictionary definition generation (LDDG), where definitions should be written using simple vocabulary.
    First, we introduce a reliable evaluation approach for DDG, based on newly proposed evaluation criteria and powered by an LLM-as-a-judge.
    To provide reference definitions for the evaluation, we construct a dataset of Japanese dictionary definitions in collaboration with a professional lexicographer.
    Validation results demonstrate that our evaluation approach agrees with human annotators at a level comparable to inter-annotator agreement.
    Second, we propose an LLM-based LDDG approach that employs iterative simplification.
    Experimental results show that our approach yields definitions that achieve high scores on the proposed criteria and exhibit high lexical simplicity.\footnote{
        Our code and dataset will be available at \url{https://github.com/Yusuke196/dictionary-gen} and \url{https://huggingface.co/datasets/yusuke196/d3j}, respectively.
    }
\end{abstract}


\begin{table*}[t!]
    \centering
    \small

    \begin{tabular}{
        >{\raggedright\arraybackslash}p{0.22\linewidth}
        p{0.66\linewidth}
    }
        \toprule
        Task & Example (Input $\rightarrow$ Output) \\
        \midrule
        Contextualized def. gen. & (\textit{``bank''}, \textit{``\ldots{} the bank of the Nile \ldots{}''}) $\rightarrow$ \textit{``the side of a river, canal, etc.''} \\
        \cmidrule(r){1-1} \cmidrule(l){2-2}
        Dictionary def. gen. (Our focus) & \textit{``bank''} $\rightarrow$ [\textit{``an organization that provides various financial services \ldots{}''}, \textit{``the side of a river, canal, etc.''}, \ldots{}] \\
        \bottomrule
    \end{tabular}

    \caption{
        Types of definition generation tasks.
        Example definitions are from \citet{lea2020advanced}.
    }
    \label{tab:def-gen-types}
\end{table*}


\section{Introduction}
\label{sec:intro}

Dictionary definitions are an essential resource for learning and understanding word senses.
They can also be used as resources for various tasks in natural language processing (NLP), e.g., word sense disambiguation.
Creating definitions for myriads of words in a dictionary, however, incurs considerable cost.

To address this issue, we study \textit{dictionary definition generation} (DDG), i.e., the generation of definitions for given headwords without context.
Specifically, we focus on definition generation for learner's dictionaries (LDs), where the definitions should consist of simple words. 
We refer to this task as \textit{learner's dictionary definition generation} (LDDG).
LDs constitute a major category of dictionaries. 
For English, six major dictionary publishers, including Oxford University Press, have produced LDs \cite{Heuberger2015}.
LDs help learners develop language capabilities, particularly in production \cite{rundell1999dictionary}.
Simple definitions, as seen in LDs, are also useful for creating vocabulary flashcards, a widely adopted resource for language learning.
However, many languages, e.g., Spanish and Japanese, lack high-quality LDs despite a large number of learners.

Recent LLM-based systems have been shown to be effective on tasks traditionally handled by human experts, e.g., Wikipedia-style article writing \cite{shao-etal-2024-assisting} and scientific paper writing \cite{lu2024ai}.
They can also simplify a wide variety of texts to improve user comprehension \cite{guidroz2025llm}, suggesting the feasibility of automatically generating correct and plain definitions that rival those written by humans. 

To evaluate generated definitions, previous studies have relied either on costly human evaluations or on metrics that do not fully account for the multi-dimensional nature of DDG evaluation.
In this paper, we propose a reproducible and fine-grained evaluation approach based on four newly designed criteria---precision, recall, sense specificity, and guideline compliance.
We evaluate them using an LLM-as-a-judge approach and a new dataset, the Dictionary Definition Dataset for Japanese (D3J), created in collaboration with a professional lexicographer.

We validate the LLM-as-a-judge approach and show that it achieves agreement with human annotators on par with inter-annotator agreement, with a Kendall's $\tau$ of 0.674 on the average score across the four criteria.
We also observe that it achieves higher agreement with humans on the criteria than conventional metrics. 

We use the evaluation approach to demonstrate that single-turn LLM prompting achieves high performance, surpassing Wiktionary on multiple criteria.
We also demonstrate that our simple LDDG approach, which combines few-shot prompting and iterative simplification, simplifies definitions without degrading their quality on other criteria.


\begin{table*}[t]
    \centering
    \small

    \begin{tabularx}{\linewidth}{
        >{\raggedright\arraybackslash}p{0.15\linewidth}
        X
    }
        \toprule
        Input & Output \\
        \toprule

        満たす, みたす, verb & ``（人の希望などを）満足させる。 \textlangle{}satisfy (e.g., someone's wishes).\textrangle{} \ldots'', ``（入れ物などを）いっぱいにする。\textlangle{}fill (e.g., a container).\textrangle{}'' \\ 
        \bottomrule
    \end{tabularx}

    \caption{
        Example input (headword followed by a reading and PoS) and output (reference) of Japanese (L)DDG.
        All data here is sourced from D3J.
        Angle-bracketed words are glosses, which are not included in the actual output; angle brackets are used in accordance with the same rule throughout the paper.
    }
    \label{tab:example-input-output}
\end{table*}


\section{Related Work}
\label{sec:related}


\subsection{Overview of Definition Generation}

\paragraph{Definition modeling and definition generation}
Studies on the generation of definitions have been referred to as \textit{definition modeling} or \textit{definition generation}, often without a clear distinction.
In contrast, we differentiate between them based on the purpose of the study.

Definition modeling is the task of generating definitions for given headwords and their embeddings in order to evaluate the embeddings \cite{noraset2017}.
The seminal work by \citet{noraset2017} studied definitions for static embeddings.
Subsequent studies, such as \citet{gadetsky-etal-2018-conditional}, generated definitions that account for the context and senses of target words.

Another line of studies has worked on generating definitions, primarily aiming to describe the meanings of unfamiliar words or phrases, with or without their embeddings.
We refer to this task as definition generation, following previous studies, such as \citet{kabiri2020, august-etal-2022-generating, huang-etal-2022-understanding}. 
Definition generation is further classified into two subcategories, illustrated in \cref{tab:def-gen-types}: (1) \textit{contextualized definition generation} (CDG), the task of generating definitions of target words given their context, and (2) \textit{dictionary definition generation} (DDG), the task of generating definitions for all senses of given (head)words.
While the present paper focuses on the latter, we review studies on both.

\paragraph{Contextualized definition generation}
Several researchers have applied definition generation techniques to describe unfamiliar words or phrases in context, aiming to support reading specialized documents (\citealp{ni-wang-2017-learning, ishiwatari-etal-2019-learning, huang-etal-2021-definition, huang-etal-2022-understanding}) or text in a foreign language \cite{zhang-etal-2023-assisting}.
An advantage of CDG is its ability to generate definitions well adapted to the context \cite{bevilacqua-etal-2020-generationary}.

\paragraph{Dictionary definition generation}
\label{para:related-ddg}
DDG requires listing all senses of a given word to enable readers to grasp the full picture of its semantics; for inputs, a part of speech (PoS) and reading (i.e., pronunciation) may also be given to specify definitions to generate.
Resources generated by DDG can also serve as lexicons for various tasks, e.g., word sense disambiguation.
Studies on DDG have been conducted in lexicography and computational linguistics. 

In lexicography, the recent emergence of ChatGPT spurred a wave of interest in computational approaches \cite{deSchryver2023}. 
\citet{lew2023chatgpt} explored the use of ChatGPT to create English monolingual dictionary entries containing sense definitions and example sentences. 
Their evaluation by human experts on 15 entries indicated that the generated definitions are comparable to those created by human lexicographers.
\citet{jakubivcek2023end} generated 99 English dictionary entries using ChatGPT and observed that ChatGPT can write correct and accessible definitions, but it sometimes repeats definitions representing virtually the same sense in different expressions, which is a phenomenon they refer to as ``false polysemy.'' 
\citet{phoodai2023} found that ChatGPT generally outperforms the Oxford Advanced Learner's Dictionary (OALD) in providing lexicographical data to English language learners.

DDG studies in computational linguistics were spearheaded by \citet{kabiri2020}, who demonstrated the utility of pre-trained multi-sense embeddings in DDG.
\citet{huang-etal-2022-jade} investigated Japanese definition generation; however, their study was based on a dataset constructed from Wikipedia and Wikidata, meaning that most of their definitions were encyclopedic descriptions of proper nouns rather than lexicographic definitions.

DDG includes LDDG as a subcategory.
Prior studies such as \citet{lew2023chatgpt} have worked on generating definitions for LDs.
However, to our knowledge, no studies have explicitly addressed LDDG as DDG with a vocabulary constraint, as we do.


\subsection{Evaluation of DDG}

The lack of a trustworthy automatic evaluation approach has been a limiting factor in previous work in DDG.
For example, the aforementioned lexicographic studies relied on human evaluation and did not establish any reproducible evaluation approaches.
The studies in computational linguistics used automatic metrics but did not verify that these metrics agree with human judgments or fully account for the arbitrariness of word sense division.

To illustrate what we call arbitrariness, take the word ``money.''
To explain its meaning, Cambridge Learner's Dictionary \cite{Cambridge2025} gives a single definition,  `the coins or banknotes (= pieces of paper with values) that are used to buy things', while OALD \cite{lea2020advanced} lists multiple definitions, such as `what you earn by working or selling things, and use to buy things' and `coins or paper notes.'
As shown here, the meaning of a given word can be represented using multiple sets of definitions with meaning divided in different ways, and thus we cannot assume that only one reference sense division exists for each headword.
In this study, we propose an evaluation approach that addresses this issue, enabling more reliable evaluation of generated definitions even when the corresponding senses are divided differently from the reference definitions.

Another line of studies, e.g., \citet{marrese-taylor-etal-2025-multilingual}, bypassed the arbitrariness issue by focusing on words with a single sense. 
In contrast, we argue that it is a critical issue for DDG, as polysemous words make up a significant proportion; for example, 40.0\% of the headwords in D3J are polysemous.
Thus, we directly address this issue.


\subsection{Resources for Japanese DDG}
\label{sec:related-resources}

This study requires a high-quality Japanese lexicographic resource for evaluation.
Some existing resources include pairs of headwords and definitions, but none are suitable.
WordNet has a Japanese version \cite{bond-etal-2009-enhancing}, but senses in WordNet are often too fine-grained \cite{strohmaier-etal-2020-secoda}.
Wiktionary has many entries for Japanese, but some of them lack definitions for essential word senses.
For example, the entry for 向き合う (`face') lacks the definition for the sense of ``\kernCJK{}（問題などに）取り組む。\kernCJK{}'' (`tackle, e.g., a problem'), as of the writing of this paper.
This leads us to create a new dataset.


\subsection{Example Sentence Generation}

To realize automated lexicography, not only DDG but also example sentence generation constitutes a key task.
Notably, \citet{cai-etal-2024-low} have shown that an LLM-based approach can generate English example sentences better than those from Oxford Dictionary.


\section{Task Formulation}
\label{sec:formulation}

We formulate the DDG task as generating a set of definitions, given a headword, PoS, and reading.
PoS is necessary; for example, it distinguishes between ある as a verb `be, exist' and ある as an adnominal `a, some.'
Reading is also necessary; for example, it distinguishes between 金 (\textit{kane}) `money' and 金 (\textit{kin}) `gold.'
Here, we provide the PoS and reading because they can help specify the definitions to generate.
For Japanese DDG, however, reading provides no additional information when the headword consists entirely of \textit{hiragana} and \textit{katakana}, which are phonographic characters; in such cases, we omit reading.
See \cref{tab:example-input-output} for an example input and output.

We define the LDDG task as DDG with a vocabulary constraint, i.e., generating definitions using only words from a given defining vocabulary.
Existing English LDs typically attempt to limit the vocabulary for writing definitions to no more than 3,500 words, resorting to words outside the vocabulary when necessary \cite{Heuberger2015}.
However, this loose constraint is more difficult to operationalize than a strict one, under which definitions must consist entirely of words in a vocabulary.
We thus employ this stricter setting while choosing a vocabulary size larger than that of typical LDs.
We investigate the appropriate vocabulary size for Japanese LDs in our dataset construction process and decide to adopt a vocabulary of approximately 16,000 words (see \cref{para:defining-vocab}).
We regard words in the defining vocabulary as simple and those out of the defining vocabulary as complex; we aim to compose definitions exclusively using simple words.


\section{Evaluation Framework}

We evaluate generated definitions using an LLM-as-a-judge approach based on Prometheus-Eval\footnote{
    \url{https://github.com/prometheus-eval/prometheus-eval}
} \cite{kim-etal-2024-prometheus}, a library using (L)LMs to perform the evaluation of natural language generation (NLG).
Our motivation here is that DDG is an NLG task, where the output space is large and no single correct answer exists \cite{gao-etal-2025-llm}.
Prometheus-Eval scores text on specified criteria, facilitating fine-grained evaluation. 
Taking as input instructions for generation, generated text, reference text, and rubrics, Prometheus-Eval prompts an evaluator LLM to produce verbal assessments followed by final scores. 
To use Prometheus-Eval for DDG, we define evaluation criteria (\cref{sec:eval-criteria}), create rubrics (\cref{sec:eval-rubrics}), and construct a dataset containing reference definitions (\cref{sec:dataset}).


\begin{figure}[t!]
    \centering
    \small
    \setlength{\tabcolsep}{0pt}
    \begin{tabularx}{\linewidth}{X}
        \toprule
        Each definition should consist of one or more sentences, that is, sequences ending with a period (句点).
        Each sentence should belong to one of the following types. \\
        1. Paraphrase: A word or phrase that shares the same or a similar meaning with the headword. Additional information---content not conveyed by the headword but helpful for understanding its meaning or usage---may be enclosed in parentheses. For example, a paraphrase of 飛ぶ \textlangle{}fly\textrangle{} could be ``\kernCJK（翼などを使って）空中を移動する。\textlangle{}Move in the air (using, e.g., wings).\textrangle{}''. Optional parts of the paraphrase may also be enclosed in parentheses. For example, a paraphrase of 上り could be ``地方や郊外から都市に向かう方向（の乗り物）。\textlangle{}Direction from rural or suburban areas toward the city or metropolis (or vehicle going in that direction).\textrangle{}''. \\
        2. Intensional: Description of the entity represented by the headword. For example, a sentence for 犬 \textlangle{}dog\textrangle{} could be ``ペットとして人間に飼われることも多い、四本の足を持つ動物。\textlangle{}A four-legged animal often kept as a pet by humans.\textrangle{}''. \\
        3. Extensional: A list of the instances or subclasses of the headword. For example, a sentence for 天気 \textlangle{}weather\textrangle{} could be ``晴れ、雨、曇りなど。\textlangle{}Sun, rain, clouds, etc.\textrangle{}''. Extensional sentences typically follow a paraphrase or intensional sentence. \\
        4. Usage: Explanation of how the headword is used, introduced by the marker, \mbox{［語法］} \textlangle{}[Usage]\textrangle{}. Use this type of sentence when it is essential for understanding the meaning or function of the headword. For example, a definition for お前 \textlangle{}you (\textit{informal})\textrangle{} could be ``あなた。［語法］自分と対等または目下の相手を指して使う。\textlangle{}You. [Usage] Used to refer to someone of equal or lower status than oneself.\textrangle{}''. \\
        \bottomrule
    \end{tabularx}

    \caption{
        Guidelines.
    }
    \label{tab:guidelines}
\end{figure}

\subsection{Criteria}
\label{sec:eval-criteria}

Our evaluation criteria correspond to four essential conditions that dictionary definitions should meet.

\paragraph{Precision and Recall}
First, definitions should describe only senses that are present in the reference; that is, precision should be high.
At the same time, they should cover all major senses in the reference; that is, recall should be high.
Our evaluation criteria thus include precision and recall.
Using these criteria enables us to handle the aforementioned issue of arbitrariness in sense division, or cases in which the headword meaning is divided differently between the system output and the reference.

\paragraph{Sense specificity}
Next, definitions should be specific; that is, they should be written so as to avoid excessive semantic overlap, including the false polysemy issue discussed in \cref{para:related-ddg}.
We refer to the criterion as sense specificity.
Note that a very vague sense is penalized on the criterion of precision instead of sense specificity, as such a sense can be viewed as not covered by the references.

\paragraph{Guideline compliance}
Finally, definitions should comply with the guidelines.
Dictionaries should present recurring elements of information in a consistent manner to avoid confusing readers \cite{atkins2008}. 
To achieve this, we created style guidelines, as described in \cref{tab:guidelines}.
They specify the types of sentences that can appear in definitions and how they should be written, providing an example for each type.
The guidelines are included in instructions to annotators and evaluator LLMs.


\subsection{Rubrics}
\label{sec:eval-rubrics}

\begin{figure*}[t!]
    \centering
    \small
    \setlength{\tabcolsep}{0pt}

    \begin{tabularx}{\linewidth}{
        X
    }
        \toprule
        \textbf{Precision:}\quad What proportion of the RESPONSE definitions correctly describe a sense covered by the REFERENCE definitions? First, read each RESPONSE definition and determine whether its sense is covered, noting that each string represents one sense and that one-to-one correspondence between the REFERENCE and RESPONSE definitions is not required. Do not consider the number of REFERENCE definitions that do not match any RESPONSE definitions. Then calculate the proportion and report it as a rounded percentage. \\
        \midrule
        \textbf{Recall:}\quad What proportion of the REFERENCE definitions have senses covered by the RESPONSE definitions? First, read each REFERENCE definition and determine whether its sense is covered, noting that each string represents one sense and that one-to-one correspondence between the REFERENCE and RESPONSE definitions is not required. Do not consider the number of RESPONSE definitions that do not match any REFERENCE definitions. Then calculate the proportion and report it as a rounded percentage. \\
        \midrule
        \textbf{Sense specificity:}\quad Do the RESPONSE definitions clearly differentiate each sense from the others? First, read all RESPONSE definitions and merge those with closely overlapping meanings, noting that each string represents one sense. For example, if the response definitions for 穴 \textlangle{}hole\textrangle{} include ``物や地面に開いた空間。\textlangle{}an empty space in an object or a surface.\textrangle{}'' and ``針などを通すための小さな開口部。\textlangle{}A small opening for passing a needle etc.\textrangle{}'', they should be merged because the senses they represent significantly overlap. Then count the number of distinct senses after merging, calculate the ratio of this number to the total number of original senses, and report it as a rounded percentage. \\
        \midrule
        \textbf{Guideline compliance:}\quad What proportion of the RESPONSE definitions comply with the guidelines described in the instructions? For each response definition, verify whether the use of parentheses and the \mbox{［語法］} \textlangle{}[Usage]\textrangle{} marker follows the guidelines. If a response definition contains the \mbox{［語法］} marker but its corresponding reference definition does not, regard it as non-compliant. Focus only on the definition style and do not consider content. After checking all response definitions, report the proportion as a rounded percentage. \\
        \bottomrule
    \end{tabularx}

    \caption{
        Rubrics for each criterion.
        ``Response definitions'' refer to output definitions to be evaluated.
    }
    \label{tab:rubrics}
\end{figure*}

For each criterion, we designed a rubric to give each headword a score from 0 to 100, as described in \cref{tab:rubrics}.
The scores represent percentages; for example, precision is measured as the proportion of generated definitions that describe a sense covered by reference definitions.
Such a design of rubrics makes scores interpretable while minimizing the subjectivity in the scoring process.
Scores from Prometheus-Eval are on a 1--5 point scale by default; thus, we modified its source code to change the scale to 0--100.
Because LLMs can introduce errors when calculating percentages, we investigate their reliability in \cref{sec:validation-analysis}.


\section{Dataset and Defining Vocabulary}
\label{sec:dataset}

To evaluate Japanese (L)DDG, we constructed a new dataset, D3J, containing definitions written in simple words.
It can be used for both LDDG and general DDG studies, because its definitions meet all requirements for general dictionaries.

Our dataset construction process involves (1) merging source data, (2) sampling headwords, (3) preparing a defining vocabulary, (4) writing definitions, and (5) reviewing them.

\paragraph{Source data}
We used three sources as bases for our dataset.
The first, JMdict \cite{breen-2004-jmdict}, is a multilingual lexical database with Japanese as the pivot language, providing well-maintained English glosses for Japanese words.
The second, \mbox{BCCWJ} (i.e., the Balanced Corpus of Contemporary Written Japanese, \citealp{maekawa2014balanced}), provides frequency information for headwords.
The third, Wiktionary, enables us to compare generated definitions with those crafted by human volunteer editors.
We access Wiktionary data using Wiktextract \cite{ylonen-2022-wiktextract}.
We sampled headwords from the intersection of these three sources.

\paragraph{Headword sampling}
We restricted our dataset to content words.
Specifically, we included nouns, verbs, \textit{i}-adjectives, \textit{na}-adjectives, adnominals (\textit{rentaishi}), and adverbs.
We did not include function words such as particles and auxiliary verbs because the qualities desired for their definitions or explanations differ from those desired for content word definitions \citep{coffey_2006_high}. 
Accordingly, our guidelines and evaluation criteria are designed for content words.

To include headwords of a variety of frequencies in our dataset, we used the BCCWJ SUW frequency list, which provides raw token counts out of the total 104.3 million.
Here, we chose SUW (i.e., short unit word) as the word unit, because the dictionary for SUW segmentation---UniDic \cite{den2007}---is relatively well-maintained and widely used for research purposes.
We defined five frequency bins---(320, 1K], (1K, 3.2K], (3.2K, 10K], (10K, 32K], and (32K, 100K]---and sampled approximately an equal number of words from each bin.
We then downsampled nouns by 70\% to balance the distribution of PoS, as nouns would otherwise vastly outnumber other PoS.

To focus on generating definitions of words important for Japanese learners, we manually excluded words that are not used in modern Japanese.
We also excluded words that are primarily used in a particular idiomatic expression because the meanings of idiomatic expressions should be interpreted or explained as a whole; for example, we excluded the word 奏する because it is almost always used in the expression 功を奏する (`prove effective') in modern Japanese.
These operations were performed by a native Japanese-speaking author with a background in NLP. 

\paragraph{Defining vocabulary}
\label{para:defining-vocab}

As discussed in \cref{sec:formulation}, the LDDG task requires a defining vocabulary to determine which words can be used in definitions; it is necessary for creating reference definitions and for performing generation.
We created our defining vocabulary by aggregating two word lists.

The first list comprises the 16,000 most frequent words from the Japanese section of TUBELEX \cite{nohejl-etal-2025-beyond}, a corpus of YouTube video subtitles.
The frequency in TUBELEX has been shown to correlate better with lexical complexity than frequencies computed with other resources, such as BCCWJ and Wikipedia \cite{nohejl-etal-2024-difficult,nohejl-etal-2025-beyond}.
We found that selecting the 16,000 most frequent words in TUBELEX allows us to write sufficiently fluent definitions within the constraint, whereas further reducing the vocabulary size makes this considerably harder.


The second list comprises ten linguistic terms that may not appear in the first list due to low frequency, but are often necessary to concisely describe meanings or usages of headwords.
For example, 口語 (`colloquial language') is included because it is a standard term used to convey that the headword is used only in ordinary conversations.\footnote{
    The other nine terms are 語法 (`word usage'), 反語 (`rhetorical expression'), 謙譲 (`modest'), 比喩 (`figure of speech'), 文語 (`literary language'), 目上 (`senior, superior'), 目下 (`junior, inferior'), 連用 (`continuative'), and 連体 (`attributive').
}

Consequently, we use a defining vocabulary consisting of just over 16,000 words, which we refer to as \textit{TUBE16K}.
Constraining the vocabulary size to this scale facilitates explaining all complex words in definitions using other entries in the same dictionary.
This is feasible because the vocabulary size of 16,000 is significantly smaller than the number of headwords found in a typical English LD; for example, OALD \cite{lea2020advanced} contains approximately 60,000 headwords.

In addition to TUBE16K, we build a smaller defining vocabulary, \textit{TUBE3K}, for lexical simplicity analysis.
We create it by aggregating the 3,000 most frequent words in TUBELEX and the aforementioned ten linguistic terms; here, 3,000 is the size of the defining vocabulary used for OALD.


\begin{figure}[t!]
    \centering
    \small
    \setlength{\tabcolsep}{0pt}

    \begin{tabularx}{\linewidth}{
        X
    }
        \toprule
        Describe the definitions of the senses of the given headword using simple words for learners of Japanese. 
        Include only senses that are expected to be familiar to most native Japanese speakers today. 
        \textit{The output MUST be a JSON array of strings, without any Markdown formatting.}\\
        \\
        \# Guidelines \\
        \{guidelines\} \\
        \bottomrule
    \end{tabularx}

    \caption{
        Instructions.
        Italicized text is presented only to LLMs.
        The \{guidelines\} placeholder is filled with the guidelines in \cref{tab:guidelines}.
    }
    \label{tab:instructions}
\end{figure}

\paragraph{Definition writing}
The aforementioned Japanese-speaking author wrote the reference definitions to satisfy the following conditions.

\begin{itemize}[leftmargin=20pt, topsep=4pt-\parskip, partopsep=0pt, itemsep=0pt, parsep=0pt]
    \item Definition text should comply with the instructions described in \cref{tab:instructions}.
    \item All words in definitions should be in TUBE16K. 
        To segment definitions into words, we use MeCab 0.996 \cite{kudo-etal-2004-applying} and UniDic 202302 \cite{den2007} throughout this paper.
        Word segmentation and simplicity checks were performed using a custom tool.
        The author took charge of writing the initial definitions in part because setting up the custom tool in the lexicographer's environment was impractical.
\end{itemize}

For reference, the author consulted various resources.
As the primary resources, they used definitions from the source data, JMdict\footnote{ 
    We incorporated English glosses from JMdict into our dataset.
    Most glosses were used as-is, but some needed modifications for the following reasons: (1) they were not appropriately divided into senses for use in monolingual Japanese dictionaries; (2) they described senses that are no longer used in modern Japanese; or (3) they lacked glosses for certain senses.
    In such cases, the Japanese-speaking author manually modified the glosses.
}
and Wiktionary, which are published under a permissive license.
In addition, they referred to several commercial dictionaries, such as Reikai Gakushu Kokugo Jiten \cite{kindaichi2022} and OALD, while avoiding verbatim copying.



\begin{table}[t!]
    \centering
    \small

    \setlength{\tabcolsep}{4pt}
    \begin{tabularx}{\linewidth}{
        l
        c
        c
        X
    }
    \toprule
     & Words & S/W & Example \\ 

    \midrule
    \textit{Frequency bin:} \\
    (320, 1K] & \phantom{0}85 & 1.27 & 偽る (`tell a lie') \\
    (1K, 3.2K] & \phantom{0}81 & 1.44 & 下旬 (`the end of the month')\\
    (3.2K, 10K] & \phantom{0}59 & 1.75 & 原理 (`principle') \\
    (10K, 32K] & \phantom{0}53 & 1.92 & 重要 (`important') \\
    (32K, 100K] & \phantom{0}47 & 2.47 & 別 (`distinction') \\

    \midrule
    Demo Set & \phantom{0}75 & 1.67 \\
    Test Set & 250 & 1.68 \\

    \midrule
    Total & 325 & 1.68 \\
    \bottomrule
    \end{tabularx}

    \caption{
        Statistics of D3J. S/W stands for ``senses per word.''
    }
    \label{tab:stats}
\end{table}


\paragraph{Definition review}
The definitions were then passed to a professional lexicographer who has worked on a major Japanese monolingual dictionary.
The lexicographer reviewed and corrected the definitions for accuracy and clarity in accordance with the instructions in \cref{tab:instructions}. 
Finally, compliance with guidelines in \cref{tab:guidelines} was checked by the Japanese-speaking author.


\paragraph{Statistics}
\cref{tab:stats} shows the statistics of D3J.
The total number of headwords is 325, and the total number of senses is 546.
The S/W column shows that frequent words tend to have more senses.
The proportion of polysemous words ranges from 23.5\% for the bin (320, 1K] to 63.8\% for (32K, 100K].
Among the 325 words, 250 are assigned to the test set and 75 to the demonstration set, whose instances can be used as few-shot examples or for model training.

We also analyze the lexical simplicity using the TUBE16K ratio and TUBE3K ratio, i.e., the ratios of the definitions that consist solely of words in TUBE16K and TUBE3K, respectively.
For the reference definitions in D3J, the TUBE16K ratio is 100\% by design; the TUBE3K ratio is 55.1\%.
For Wiktionary definitions in D3J, the TUBE16K ratio is 65.6\% and the TUBE3K ratio is 21.1\%. 
This result indicates that our reference definitions are lexically simpler than those in Wiktionary.





\section{Validating Evaluation Approach}
\label{sec:eval-validation}

To validate our LLM-as-a-judge approach, we measure its agreement with human annotators using Kendall's $\tau$.
Kendall's $\tau$ measures the degree of association between two variables, even when they are derived from different measures, such as our annotators' scores and a BLEU score.
It ranges from \textminus{}1 to 1, where \textminus{}1 represents complete disagreement and 1 perfect agreement.


\begin{table*}[t!]
    \centering
    \small

    \begin{tabular}{lccccc}
    \toprule
      & Overall & Precision & Recall & Sense Specificity & Guideline Compliance \\

    \midrule
    Claude Sonnet 4.6--human & 0.556 & 0.483 & 0.256 & \textbf{0.468} & \textbf{0.919} \\
    GPT 5.5--human & \textbf{0.674} & \textbf{0.563} & \textbf{0.539} & \textbf{0.468} & \textbf{0.919} \\

    \cmidrule(r){1-1} \cmidrule(l){2-6}
    Concat BLEU--human & 0.312 & 0.097 & 0.040 & 0.269 & 0.111 \\
    Match BLEU--human & 0.240 & 0.015 & 0.073 & 0.253 & 0.114 \\
    Concat BERTScore F1--human & 0.329 & 0.151 & 0.078 & 0.205 & 0.182 \\
    Concat BERTScore P--human & 0.310 & 0.160 & \tableminus{}0.073 & 0.205 & 0.203 \\
    Concat BERTScore R--human & 0.230 & \tableminus{}0.021 & 0.190 & 0.227 & 0.107 \\
    Match BERTScore F1--human & 0.203 & \tableminus{}0.001 &  0.134 & 0.206 & 0.142 \\
    Match BERTScore P--human  & 0.212 &  0.083 &  0.001 & 0.163 & 0.179 \\
    Match BERTScore R--human  & 0.114 & \tableminus{}0.113 &  0.188 & 0.209 & 0.048 \\

    \cmidrule(r){1-1} \cmidrule(l){2-6}
    Inter-human & 0.577 & 0.533 & 0.465 & 0.532 & 0.838 \\
    \bottomrule
    \end{tabular}

    \caption{
        Machine--human and inter-human agreement on evaluation scores, as Kendall's $\tau$.
        The bold font denotes the highest machine--human agreement value.
    }
    \label{tab:validation-tau}
\end{table*}


\subsection{Annotation}

We sampled a subset of 100 headwords for validation from the D3J test set.
We generated definitions for the headwords by zero-shot prompting with GPT-4o (\texttt{gpt-\allowbreak{}4o-\allowbreak{}2024-\allowbreak{}08-\allowbreak{}06}). 
For the prompt, we used the instructions in \cref{tab:instructions}.

We recruited four native Japanese speakers with NLP backgrounds and asked them to score definitions for each headword across the four criteria, providing the guidelines and rubrics discussed in Sections \ref{sec:eval-criteria} and \ref{sec:eval-rubrics}.
After annotating the first 15 headwords, they received feedback intended to correct any misunderstandings about the task.
The annotators were also asked to report any problematic reference definitions, e.g., definitions missing widely used senses, because such references hinder appropriate evaluation.
In this manner, we identified four problematic definitions and excluded them from the validation subset. 
However, we corrected them and included them in the main dataset to maintain its size.

Finally, we excluded one annotator as an outlier because their agreement with the other annotators on the overall score was the lowest, where the overall score is calculated as the average of the four criteria.\footnote{
    Their low agreement with the other annotators suggests that the feedback on 15 headwords was insufficient for this task.
    One possible solution to this issue is to conduct a screening process in which we exclude one or two annotators who show low agreement with the gold annotations we prepare.
}


\subsection{Models and Baselines}
\label{sec:models-and-baselines}

For evaluator LLMs, we experiment with two proprietary LLMs: GPT-5.5 (\texttt{gpt-\allowbreak{}5.5-\allowbreak{}2026-\allowbreak{}04-\allowbreak{}23}) and Claude Sonnet 4.6.
Models from the GPT and Claude families ranked high on a leaderboard for Japanese tasks, the Nejumi Leaderboard 4\footnote{\url{https://wandb.ai/llm-leaderboard/nejumi-leaderboard4/reports/Nejumi-LLM-4--VmlldzoxMzc1OTk1MA}} at the time of model selection.

We also compute the agreement between conventional metrics based on BLEU or BERTScore and human annotators.
Each metric produces a single score for each set of definitions.
We use this score as the metric's score for each evaluation criterion to compute its agreement with the annotators.

\paragraph{BLEU} We use BLEU \cite{papineni2002}, the de facto standard metric for DDG and definition modeling (\citealp{noraset2017,kabiri2020}; \textit{inter alia}). 
For implementation, we use the sacrebleu package \cite{post-2018-call}. 

\paragraph{BERTScore} We use BERTScore \cite{zhang2020}, which was applied to DDG by \citet{bevilacqua-etal-2020-generationary}. For implementation, we use the bert-score package.\footnote{
    \url{https://pypi.org/project/bert-score/}
}

\paragraph{} We experiment with two approaches to aggregating the scores for a set of definitions:

\paragraph{Concat} The concatenation-based approach computes BLEU (a single metric) or BERTScore (P, R, and F1) for a headword by applying the metrics to a reference string and a hypothesis string built by concatenating definitions using ``／'' (full-width slash) as the separator.
One potential issue with this approach is that the scores are influenced by the order of definitions.\footnote{
    Note that this does not affect BLEU or BERTScore for definitions of words disambiguated by context \citep{bevilacqua-etal-2020-generationary} or monosemous words \citep{august-etal-2022-generating}.
}

\paragraph{Match} The matching-based approach is order-agnostic and relies on best-matching definitions.
It computes precision and recall (or a single precision-based score in the case of BLEU) for each hypothesis-reference pair of definitions and then aggregates the scores using the same logic that BERTScore uses at the token level. 
See \cref{sec:appendix-baselines} for the algorithms.

\subsection{Results}
\label{sec:validation-results}

Inter-human agreement, computed as the mean $\tau$ of three annotator pairs, is shown at the bottom of \cref{tab:validation-tau}.
$\tau$ values are at least 0.465 across the four criteria.
Although this level of agreement may seem low, some degree of disagreement is common in lexical semantics.
For example, \citet{mihalcea-etal-2004-senseval} constructed a dataset for the English lexical sample task in Senseval-3 and reported a macro-$\kappa$ inter-annotator agreement of 0.35. 
\citet{jurgens-2013-embracing} measured agreement between annotators to investigate several methods to collect high-quality annotations for the word sense disambiguation task.
Their experiments show that even the method achieving the highest inter-annotator agreement yielded Krippendorff's $\alpha$ of only 0.220--0.511. 
While these studies employed a different metric from ours, their metrics range from \textminus{}1 to 1 like $\tau$. 
When computed as the mean Krippendorff's $\alpha$ across the three pairs, our inter-human agreement values range from 0.459 to 0.821, which are comparable to or higher than those reported by \citet{jurgens-2013-embracing}.

\cref{tab:validation-tau} also shows the machine--human agreement, which is computed as the average $\tau$ across three pairs consisting of an automated metric and an annotator.
GPT-5.5 achieves the highest agreement with humans, surpassing the level of inter-human agreement on precision, recall, and guideline compliance.
GPT-5.5 also achieves substantially higher agreement than the BERTScore and BLEU metrics.
We should note that BERTScore and BLEU were not designed to evaluate the criteria we defined.
Nonetheless, our approach achieved higher agreement across all criteria and on the overall score.
Given these results, we use GPT-5.5 as our evaluator LLM.


\begin{figure}[t]
    \centering
    \small

    \setlength{\tabcolsep}{0pt}
    \begin{tabularx}{\linewidth}{
        X
    }
        \toprule
        Both response definitions correctly describe senses covered by the reference. The first definition, ``（物が）軽く動いて安定しない状態になる。'' corresponds to the physical sense ``揺れる。\textlangle{}Shake.\textrangle{}'' The second definition, ``（心や考えが）不安定になる。\textlangle{}(Of mind or thoughts) become unstable.\textrangle{}'' corresponds to the abstract sense ``（物事や気持ちなどが）不安定になる。\textlangle{}(Of things or feelings) become unstable.\textrangle{}'' Since all 2 response definitions match senses in the reference, the proportion is 2/2. $[$RESULT$]$ 100 \\

         \bottomrule
    \end{tabularx}

    \caption{
        Verbal assessment by GPT-5.5 for 揺らぐ (`shake, sway') on precision.
        See \cref{sec:appendix-eval-results} for corresponding reference definitions, generated definitions, and assessments on all four criteria.
    }
    \label{tab:assessment-example}
\end{figure}

\subsection{Analysis}
\label{sec:validation-analysis}

We also investigated whether making LLMs calculate final scores leads to errors.
That is, we focused here on the consistency between the initial assessments and the final scores, rather than on the correctness of the final scores.
We sampled 15 headwords and manually checked whether the final score for each of the four criteria by GPT-5.5 was appropriately derived or calculated from the preceding assessments.
We did not observe any inconsistencies in the 60 samples.

\cref{tab:assessment-example} shows an example of the verbal assessment and final score for 揺らぐ (`shake, sway') on the precision criterion.
Here we find that our approach evaluates definition quality as instructed in the rubrics.

The scores do not differ markedly between our LLM evaluator and human annotators, although the LLM tends to give slightly lower scores.
The mean overall score of GPT-5.5 is 87.0, and the score range is [58.2, 100.0].
The mean overall scores of the three annotators are 87.1, 89.0, and 88.5, and their score ranges are [50.0, 100.0], [62.5, 100.0], and [50.0, 100.0], respectively.


\section{LDDG Approach}
\label{sec:approach}


\subsection{Single-turn Prompting}

First, we investigate single-turn, i.e., zero-shot and few-shot, prompting.
We provide LLMs with the instructions in \cref{tab:instructions}, which were also used to construct the dataset.
In the few-shot setting, we randomly sample 5 pairs of an input (headword, reading, and PoS) and an output (definition set) from the demonstration set and include them in the instructions.

We use two proprietary LLMs---GPT-5.5 and Claude Sonnet 4.6---and two open-weight LLMs---Qwen3.6-27B \cite{yang2025qwen3} and Llama-3.3-Swallow-70B-Instruct-v0.4 \cite{Fujii:COLM2024} (hereafter Qwen and Swallow, respectively).
They were among the highly ranked LLMs on the Nejumi Leaderboard 4 at the time of model selection. 
Swallow is continually pre-trained on Japanese data.

For inference, we use the default reasoning setting for each model, except Qwen, for which we disabled reasoning because it resulted in more invalid JSON outputs. 
The temperature for text generation was set at 0.0.
Generation was constrained to produce an array of strings.
We treated outputs that did not comply with this constraint as empty arrays. 
For open-weight LLMs, we perform 4-bit quantization using bitsandbytes.\footnote{
    \url{https://github.com/TimDettmers/bitsandbytes}
}

\subsection{Iterative Simplification}

Second, we simplify the outputs of single-turn prompting via an LLM-based iterative simplification approach inspired by \citet{oshika-etal-2024-simplifying}, which we refer to as \textit{IterSim}.
Whereas single-turn prompting only instructs LLMs to write definitions using simple words, IterSim explicitly attempts to eliminate all complex words.

The overview of the IterSim algorithm is given in \cref{alg:itersim}.
We first identify complex words in the initial definitions by performing word segmentation and extracting those not in TUBE16K (line 1).
We then try to remove complex words one by one while ensuring that the revised version remains an accurate definition of the headword (lines 2--8).

\Call{Simplify}{{}} on line 4 is detailed in \cref{alg:simplify}.
It performs zero-shot prompting up to two times using the same LLM as in DDG and the prompt in \cref{tab:simplify-prompt}:
(1) Prompt the LLM to simplify the word \Var{w} in \Var{def} (line 6).
(2) If the number of complex words decreases in Step (1), return \Var{True} for \Var{isSuccess} and the output for \Var{simDef} (line 8).
(3) If new complex words are introduced, attempt to simplify \Var{w} in \Var{def} again, this time banning the new complex words via banned\_words in the prompt (line 12--13, 6).
(4) If there are no new complex words, but \Var{w} remains in the output, attempt to simplify \Var{w} in the output; this step handles cases when the LLM simplifies a different word than expected (line 14--15, 6).
(5) If the number of complex words does not decrease after the second attempt, return \Var{False} for \Var{isSuccess} (line 19).



\begin{algorithm}[t]
    \small
    \caption{IterSim}
    \label{alg:itersim}

    \begin{algorithmic}[1] 
        \Input{Headword \Var{h}, definition \Var{def}}
        \Output{Simplified definition \Var{currDef}}

        \State $\BVar{compWords} \leftarrow \Call{FindComplexWords}{\Var{def}}$
        \State $\Var{currDef} \leftarrow \Var{def}$

        \ForAll{$\Var{w} \in \BVar{compWords}$}
            \State $\Var{isSuccess}, \Var{simDef} \leftarrow \Call{Simplify}{\Var{currDef}, \Var{w}, \Var{h}}$
            \If{\Var{isSuccess}}
                \State $\Var{currDef} \leftarrow \Var{simDef}$
            \EndIf
        \EndFor

        \State \textbf{return} \Var{currDef}

    \end{algorithmic}
\end{algorithm}

\begin{algorithm}[t]
    \small\raggedright
    \caption{Simplification targeting the word \Var{w}}
    \label{alg:simplify}

    \begin{algorithmic}[1] 
        \Input{Definition \Var{def}, complex word \Var{w}, headword \Var{h}}
        \Output{Success flag \Var{isSuccess}, simplified definition \Var{simDef}}

        \State $\Var{inferenceCount} \gets 0$
        \State $\BVar{cWords} \gets \Call{FindComplexWords}{\Var{def}} $
        \State $\Var{cCount} \gets \Call{CountComplexWords}{\Var{def}}$
        \State $\BVar{banned} \gets []$

        \While{$\Var{inferenceCount} < 2$}
            \State $\Var{simDef} \gets \Call{Inference}{\Var{h}, \Var{def}, \Var{w}, \BVar{banned}}$

            \If{$\Call{CountComplexWords}{\Var{simDef}} < \Var{cCount}$}
                \State \Return $\Var{True}, \Var{simDef}$
            \EndIf

            \State $\BVar{cWords2} \gets \Call{FindComplexWords}{\Var{simDef}} $
            \State $\BVar{newCWords} \gets \Call{Set}{\BVar{cWords2}} - \Call{Set}{\BVar{cWords}}$

            \If{$\Call{Length}{\BVar{newCWords}} > 0$}
                \State $\BVar{banned} \gets \BVar{banned} + \BVar{newCWords}$
            \Else
                \State $\Var{def} \gets \Var{simDef}$
            \EndIf

            \State $\Var{inferenceCount} \gets \Var{inferenceCount} + 1$
        \EndWhile

        \State \Return $\Var{False}, \varepsilon$ 
    \end{algorithmic}
\end{algorithm}

\begin{figure}[t]
    \centering
    \small

    \setlength{\tabcolsep}{0pt}
    \begin{tabularx}{\linewidth}{
        X
    }
        \toprule
        The given definition contains a complex word that could be difficult for learners. Rewrite it without using the complex word, ensuring that (1) the revised version remains an accurate and fluent representation of the headword's sense and (2) the revised version does not contain the banned words if any are provided. [...] \\
        \\
        Headword: \{headword\} \\
        Definition: \{definition\} \\
        Target word: \{target\_word\} \\
        Banned words: \{banned\_words\} \\
        Simplified Definition: \\
        \bottomrule
    \end{tabularx}

    \caption{
        Prompt for simplification.
        Curly braces denote a placeholder.
    }
    \label{tab:simplify-prompt}
\end{figure}



\begin{table*}[t!]
    \centering
    \small

    \begin{tabular}{llcccccc}
    \toprule
     & & Overall & Precision & Recall & Specificity & Compliance & TUBE16K \\

    \midrule
    \multirow[t]{4}{*}{Zero-shot} & Claude Sonnet 4.6 & 89.0 & 90.5 & 95.5 & 92.0 & 78.3 & 85.1 \\
     & GPT-5.5 & 90.5 & 85.0 & 98.6 & 84.8 & 93.5 & 92.7 \\
     & Qwen & 78.8 & 82.7 & 86.4 & 65.6 & 80.6 & 84.8 \\
     & Swallow & 68.9 & 74.5 & 66.4 & 63.2 & 71.7 & 92.3 \\
    \midrule
    \multirow[t]{4}{*}{Few-shot} & Claude Sonnet 4.6 & 93.5{\scriptsize\textpm 0.7} & 92.1{\scriptsize\textpm 1.8} & 93.5{\scriptsize\textpm0.9} & 93.3{\scriptsize\textpm 1.6} & 95.2{\scriptsize\textpm 1.2} & 91.4{\scriptsize\textpm 1.6} \\
     & GPT-5.5 & 91.7{\scriptsize\textpm 0.3} & 87.4{\scriptsize\textpm 0.3} & 98.4{\scriptsize\textpm 0.2} & 86.2{\scriptsize\textpm 1.0} &94.7{\scriptsize\textpm 0.5} & 91.3{\scriptsize\textpm 0.0} \\
     & Qwen & 84.9{\scriptsize\textpm 2.4} & 87.5{\scriptsize\textpm 4.4} & 78.3{\scriptsize\textpm 3.3} & 78.2{\scriptsize\textpm 5.3} & 95.6{\scriptsize\textpm 1.5} & 91.4{\scriptsize\textpm 0.4} \\
     & Swallow & 74.9{\scriptsize\textpm 2.6} & 66.1{\scriptsize\textpm 4.0} & 59.2{\scriptsize\textpm 1.6} & 81.6{\scriptsize\textpm 2.7} &92.7{\scriptsize\textpm 2.9} & 92.7{\scriptsize\textpm 1.6} \\

    \midrule
    Wiktionary &  & --- & 72.3 & 94.8 & 80.7 & --- & 65.6 \\
    \bottomrule
    \end{tabular}

    \caption{
        Results of single-turn prompting, as mean scores across definition sets.
        Few-shot scores are averaged over three runs, where randomness arises from example selection.
        \textpm{} denotes standard deviation.
        The Wiktionary score on guideline compliance is not applicable, as its editors do not see our guidelines while writing definitions.
    }
    \label{tab:baseline-results}

    \vspace{1.2\baselineskip}
    \setlength{\tabcolsep}{4pt} 
    \begin{tabularx}{\linewidth}{
        >{\raggedright\arraybackslash}p{\widthof{Distress\textrangle}}
        >{\raggedright\arraybackslash}p{0.265\linewidth}
        >{\raggedright\arraybackslash}p{0.51\linewidth}
        >{\raggedright\arraybackslash}X
    }
        \toprule
        Word & Reference & Output & Note \\
        \midrule
        揺らぐ \textlangle{}Shake/\allowbreak{}Waver\textrangle{} & 
            ``揺れる。\textlangle{}Shake.\textrangle{}'', ``\kernCJK{}（物事や気持ちなどが）不安定になる。\kernCJK{} \textlangle{}(Of things, feelings, etc.) become unstable.\textrangle{}'' & 
            ``ゆっくりと左右や前後に動く。\textlangle{}Slowly move side to side or back and forth.\textrangle{}'', ``\kernCJK{}（気持ちや考えなどが）不安定になる。\textlangle{}(Of feelings, thoughts, etc.) become unstable.\textrangle{}'' & 
            Full score on all criteria. \\
        \cmidrule(r){1-1} \cmidrule(l){2-4}
        大きい \textlangle{}Big/\allowbreak{}Loud\textrangle{} & 
            ``（物の高さや幅、広さ、程度、数などが）普通を超えている。\textlangle{}(Of height, width, size, degree, number, etc.) exceeding normal.\textrangle{}'', \ldots & 
            ``（物の形・面積・体積などが）広い、または多くの空間を占めている。\textlangle{}(Of shape, area, volume, etc.) wide or occupying a large amount of space.\textrangle{}'', \ldots, ``（人の心・態度などが）細かいことを気にしない。\textlangle{}(Of mind, attitude, etc.) not worrying about small things.\textrangle{}'' &
            Low precision. \\
        \bottomrule
    \end{tabularx}

    \caption{
        Example outputs of few-shot prompting with Claude Sonnet 4.6.
    }
    \label{tab:generation-examples-fewshot}

    \vspace{1.2\baselineskip}
    \begin{tabular}{llccccc}
    \toprule
     & Overall & Precision & Recall & Specificity & Compliance & TUBE16K \\
    \midrule
     Few-shot & 93.5{\scriptsize\textpm 0.7} & 92.1{\scriptsize\textpm 1.8} & 93.5{\scriptsize\textpm 0.9} & 93.3{\scriptsize\textpm 1.6} & 95.2{\scriptsize\textpm 1.2} & 91.4{\scriptsize\textpm 1.6} \\
     Few-shot + IterSim & 93.6{\scriptsize\textpm 0.7} & 92.1{\scriptsize\textpm 1.9} & 93.6{\scriptsize\textpm 0.8} & 93.5{\scriptsize\textpm 1.6} & 95.3{\scriptsize\textpm 1.2} & 98.9{\scriptsize\textpm 0.4} \\
    \bottomrule
    \end{tabular}

    \caption{
        Results of IterSim. Claude Sonnet 4.6 is used for all approaches. \textpm{} denotes standard deviation.
    }
    \label{tab:itersim}

    \vspace{1.2\baselineskip}
    \setlength{\tabcolsep}{4pt}
    \begin{tabularx}{\linewidth}{
        >{\raggedright\arraybackslash}p{\widthof{\textlangle{}Ances-}}
        >{\raggedright\arraybackslash}p{0.27\linewidth}
        >{\raggedright\arraybackslash}X
    }
        \toprule
        Word & Reference & Output (Before IterSim $\rightarrow$ After IterSim) \\

        \midrule
        布 \textlangle{}Clo\-th\textrangle{} &
            ``糸などをもとに作った、服を作るためなどに使う薄い材料。\textlangle{}Thin material made from threads, used for clothing and similar items.\textrangle{}'' &
            ``糸を縦と横に\textbf{織っ}て作った、薄くて平らな素材。\textlangle{}Thin and flat material made by \textbf{weaving} threads vertically and horizontally.\textrangle{}'' \newline 
            $\rightarrow$ ``糸を縦と横に組み合わせて作った、薄くて平らな素材。\textlangle{}Thin and flat material made by combining threads vertically and horizontally.\textrangle{}'' \\
        
        \bottomrule
    \end{tabularx}

    \caption{
        Example outputs of few-shot prompting and IterSim.
        Bold font denotes complex words.
    }
    \label{tab:examples-itersim}
\end{table*}

\section{Experimental Results}


\subsection{Single-turn Prompting}

\cref{tab:baseline-results} shows the results of single-turn prompting.
Few-shot prompting improves the overall score over zero-shot prompting for all models.
Proprietary LLMs outperform open-weight LLMs for both zero-shot and few-shot prompting.
The best-performing combination is few-shot prompting with Claude Sonnet 4.6; it achieves scores of at least 92.1 on all criteria, outperforming few-shot prompting with GPT-5.5 on precision, sense specificity, and guideline compliance.\footnote{
    Among the proprietary models, Claude Sonnet 4.6 performed better on definition generation, while GPT-5.5 performed better on definition evaluation.
    One possible explanation is that the two tasks differ in nature: evaluation relies on rubrics and reference definitions, whereas generation uses guidelines and optional few-shot examples.
}
Moreover, it outperforms Wiktionary on precision and sense specificity by 10--20 points, while achieving comparable performance on recall.
\cref{tab:generation-examples-fewshot} shows a successful example by the best combination, where the output definitions correctly describe the senses of 揺らぐ (`shake, sway').

Nonetheless, some issues remain.
For instance, definitions for some words suffer from low precision.
\cref{tab:generation-examples-fewshot} shows the output definitions for 大きい (`big').
Its second output definition (``\ldots{}細かいことを気にしない。\kernCJK{}'') is neither present in the reference nor attested as a modern sense of the word.\footnote{
    心が広い (literally ``having a broad heart'') means ``not worrying about small things,'' so the model may have confused 大きい (`big') and 広い (`broad').
}
Although such errors are rare, human review is desirable to ensure definition quality.

A single-variable correlation analysis (\cref{tab:pcc-overall} in \cref{sec:appendix-analysis}) shows that while the effect of PoS is inconsistent across models, both frequent and polysemous words are consistently associated with lower overall scores. 
Although polysemy is generally associated with high frequency and its ratio varies by PoS (see \cref{tab:polysemy-freq-by-pos}), we show that polysemy has a large negative effect on recall even when controlling for PoS and frequency (see \cref{tab:regression-analysis}).
The models tend to generate more senses compared to the reference data (see \cref{tab:polysemy-stats}). 
For GPT-5.5, the tendency is particularly strong and persists even in the few-shot setting, contributing to its high recall and relatively low precision.

We assessed simplicity using the TUBE16K ratio (see \cref{sec:dataset}) of the generated definitions. 
The 91.4\% ratio achieved by few-shot prompting with Claude Sonnet 4.6 indicates that some definitions contain complex words.
We explore how to bring it closer to 100\% in the following section.


\subsection{Iterative Simplification}

\cref{tab:itersim} reports the performance of IterSim, provided with initial definitions generated via few-shot prompting with Claude Sonnet 4.6; it shows scores before and after applying IterSim.
We find that applying IterSim increases the TUBE16K ratio to nearly 100\% while keeping the other scores largely unchanged, suggesting that IterSim can simplify definitions without sacrificing their essential qualities.
See \cref{tab:examples-itersim} for example outputs.
The definition for 布 (`cloth') was successfully simplified, with the complex word 織る (`weave') being rephrased while the meaning of the definition as a whole was preserved.


\section{Conclusion}

In this paper, we designed new evaluation criteria for DDG, proposed an LLM-as-a-judge approach based on them, and created an LDDG dataset in collaboration with a professional lexicographer.
We used this evaluation approach and dataset to demonstrate that single-turn LLM prompting generates definitions that achieve strong scores (at least 92.1 out of 100.0) across four criteria, outperforming Wiktionary definitions in precision and sense specificity.
We also showed that our iterative simplification method, IterSim, can simplify the definitions to satisfy the vocabulary constraint without degrading their quality on the other criteria.
These results suggest that our LLM-based approach has practical potential for lexicography, particularly for learner's dictionaries.

Although we addressed Japanese LDDG, many of our contributions, such as the criteria and the LLM-as-a-judge evaluation approach, are potentially applicable to DDG studies for any language and any type of dictionary.
We believe that this study can serve as an important step towards automated lexicography in general.


\section{Limitations}
\label{sec:limitations}

The size of our dataset, 325 headwords (546 definitions), may seem modest. 
It is not particularly small, however, for an evaluation dataset with a relatively high cost of construction.
For example, the lexical simplification dataset by \citet{uchida-etal-2018-cefr} contains 406 target words.
Our dataset was created in collaboration with a professional lexicographer, and it contains rich information for each headword.
Because its construction requires substantial effort and cost, it was difficult to obtain more data.

This study focused on definitions of content words, but learner's dictionaries also provide definitions of function words, addressing grammar and pragmatics.
Function words, particularly those that primarily fulfill grammatical roles, require different criteria for definition evaluation.
In addition, we did not address definition generation for multiword expressions (MWEs), that is, idiomatic sequences of multiple words \cite{BaldwinMWETextbook}.
MWEs include verbal MWEs, nominal MWEs, and functional expressions.
Functional expressions require different evaluation criteria, as do function words.
Addressing these types of items remains an issue for future research.




\section*{Acknowledgments}

We thank Professor Makoto Yamazaki for his assistance in constructing the D3J dataset.
We are grateful for the insightful comments by anonymous reviewers.
Financially, this work was supported by JST SPRING Grant Number JPMJSP2140.


\bibliography{anthology,custom}
\bibliographystyle{acl_natbib}


\appendix

\section{Validating Evaluation Approach}
\label{sec:appendix-eval-validation}

\subsection{Baselines}
\label{sec:appendix-baselines}

\begin{algorithm}[ht]
    \small
    \caption{Matching-based BERTScore}
    \label{alg:bertscore}

    \begin{algorithmic}[1] 
        \Input{
            Reference definitions $\BVar{ref}$,
            hypothesis definitions $\BVar{hyp}$
            }
        \Output{Precision $P$, recall $R$, F1 score $F_1$}

        \For{$i \textbf{ in } (1, \ldots, \Call{Length}{\BVar{ref}})$}
            \For{$j \textbf{ in } (1, \ldots, \Call{Length}{\BVar{hyp}})$}
                \State $p_{i,j} \gets \Call{BERTScoreP}{\BVar{ref}[i], \BVar{hyp}[j]}$
                \State $r_{i,j} \gets \Call{BERTScoreR}{\BVar{ref}[i], \BVar{hyp}[j]}$
            \EndFor
        \EndFor

        \State $P \gets \sum_{j=1}^{|\BVar{hyp}|} \left(\max_{i=1}^{|\BVar{ref}|} p_{i,j}\right) /\,\Call{Length}{\BVar{hyp}}$

        \State $R \gets \sum_{i=1}^{|\BVar{ref}|} \left(\max_{j=1}^{|\BVar{hyp}|} r_{i,j}\right) /\,\Call{Length}{\BVar{ref}}$

        \State $F_1 \gets 2PR/(P + R)$

        \State \Return $P$, $R$, $F_1$
    \end{algorithmic}
\end{algorithm}

\begin{algorithm}[ht]
    \small
    \caption{Matching-based BLEU}
    \label{alg:bleu}

    \begin{algorithmic}[1] 
        \Input{
            Reference definitions $\BVar{ref}$,
            hypothesis definitions $\BVar{hyp}$
            }
        \Output{BLEU score $P$ (precision-based)}

        \For{$i \textbf{ in } (1, \ldots, \Call{Length}{\BVar{ref}})$}
            \For{$j \textbf{ in } (1, \ldots, \Call{Length}{\BVar{hyp}})$}
                \State $p_{i,j} \gets \Call{BLEU}{\BVar{ref}[i], \BVar{hyp}[j]}$
            \EndFor
        \EndFor

        \State $P \gets \sum_{j=1}^{|\BVar{hyp}|} \left(\max_{i=1}^{|\BVar{ref}|} p_{i,j}\right) /\,\Call{Length}{\BVar{hyp}}$

        \State \Return $P$
    \end{algorithmic}
\end{algorithm}

\cref{alg:bertscore} and \cref{alg:bleu} describe analogous algorithms for matching-based BERTScore and BLEU, respectively.

\subsection{Results}
\label{sec:appendix-eval-results}

\begin{table*}[ht]
    \centering
    \small
    \setlength{\tabcolsep}{4pt}

    \begin{tabularx}{\linewidth}{
        lX
    }
        \toprule
        Reference & ``揺れる。\kernCJK{} \textlangle{}Shake or swing.\textrangle{}'', ``\kernCJK{}（物事や気持ちなどが）不安定になる。\kernCJK{} \textlangle{}(Of things, feelings, etc.) become unstable.\textrangle{}'' \\

        \midrule
        Hypothesis & ``\kernCJK{}（物が）軽く動いて安定しない状態になる。\textlangle{}(Of things) lightly move and become unstable.\textrangle{}'', ``\kernCJK{}（心や考えなどが）不安定になる。\textlangle{}(Of mind, thoughts, etc.) become unstable.\textrangle{}'' \\

        \midrule
        Assessment 
          & \textbf{Precision:}\quad Both response definitions correctly describe senses covered by the reference. The first definition, ``（物が）軽く動いて安定しない状態になる。\textlangle{}(Of things) lightly move and become unstable.\textrangle{}'' corresponds to the physical sense ``揺れる。\textlangle{}Shake.\textrangle{}'' The second definition, ``（心や考えが）不安定になる。\textlangle{}(Of mind, thoughts, etc.) become unstable.\textrangle{}'' corresponds to the abstract sense ``（物事や気持ちなどが）不安定になる。\textlangle{}(Of things or feelings) become unstable.\textrangle{}'' Since all 2 response definitions match senses in the reference, the proportion is 2/2. $[$RESULT$]$ 100 \\
          & \textbf{Recall:}\quad The response covers both reference senses. The first definition covers the physical sense of ``揺れる \textlangle{}shake\textrangle{}'' by describing an object moving lightly and becoming unstable. The second definition covers the abstract sense of becoming unstable in feelings or thoughts, corresponding to the reference sense ``物事や気持ちなどが不安定になる \textlangle{}things, feelings, etc. become unstable\textrangle{},'' though it is slightly narrower because it does not explicitly mention ``物事 \textlangle{}things\textrangle{}.'' Since both reference senses are covered, the coverage proportion is 2/2. [RESULT] 100 \\
          & \textbf{Sense specificity:} The response provides two definitions and they represent two distinct senses: one for physical objects moving/shaking and becoming unstable, and one for mental states or thoughts becoming unstable. Although both mention instability, the domains are clearly separated, so they should not be merged. Thus, 2 distinct senses remain out of 2 original response definitions, giving a rounded percentage of 100\%. [RESULT] 100 \\
          & \textbf{Guideline compliance:} 両方の定義が、句点で終わる文として書かれており、JSON配列の文字列形式にも合っています。\textlangle{}The two definitions are written as sentences ending with a period, and also conform to the JSON array string format.\textrangle{} 1つ目の「（物が）軽く動いて安定しない状態になる。」は、括弧を使って対象を補足しており、ガイドライン上許容される形です。\textlangle{}The first definition, "（物が）軽く動いて安定しない状態になる。", uses parentheses to supplement the subject, which is allowed by the guidelines.\textrangle{} 2つ目の「（心や考えが）不安定になる。」も、括弧内で対象を示す自然な補足になっており、形式上問題ありません。\textlangle{}The second definition, "（心や考えが）不安定になる。", uses parentheses to supplement the subject, which is allowed by the guidelines.\textrangle{} ［語法］マーカーは使われておらず、参照定義との不一致もありません。\textlangle{}The [語法] marker is not used, and there is no inconsistency with the reference definition.\textrangle{} したがって、2件中2件がガイドラインに準拠しています。\textlangle{}Therefore, all 2 definitions are compliant with the guidelines.\textrangle{} [RESULT] 100 \\

        \bottomrule
    \end{tabularx}

    \caption{Verbal assessment by GPT-5.5 for 揺らぐ (`shake, sway').}
    \label{tab:assessment-example-full}
\end{table*}

\cref{tab:assessment-example-full} describes the full assessment for an example headword by GPT-5.5.


\FloatBarrier

\section{Analysis of Results}
\label{sec:appendix-analysis}

\cref{tab:pcc-overall} shows the correlations of overall zero-shot model performance with log frequency, polysemy, and each PoS. \cref{tab:polysemy-freq-by-pos} shows the ratios of polysemous words and frequency statistics by PoS in the test set of D3J, on which we performed further analysis. 
\cref{tab:regression-analysis} shows the regression analysis of zero-shot model performance with the unique contributions of log frequency, polysemy, and PoS to the explained variance. \cref{tab:polysemy-stats} compares polysemy statistics of the reference data, the model outputs, and Wiktionary.

\begin{table*}[htbp]
  \centering\small
  \setlength{\tabcolsep}{5pt}
\begin{tabular}{lccccccc}
\toprule
Model & Frequency & Polysemy & Adverb & \textit{i}-adjective & \textit{na}-adjective & Noun & Verb \\
\midrule
Claude Sonnet 4.6
& \niceminus{}0.298$^{***}$
& \niceminus{}0.194$^{**\phantom{*}}$
& \niceminus{}0.268$^{***}$
& \nominus{}0.005$^{\phantom{***}}$
& \nominus{}0.005$^{\phantom{***}}$
& \nominus{}0.061$^{\phantom{***}}$
& \nominus{}0.078$^{\phantom{***}}$ \\

GPT-5.5
& \niceminus{}0.278$^{***}$
& \niceminus{}0.003$^{\phantom{***}}$
& \niceminus{}0.053$^{\phantom{***}}$
& \niceminus{}0.117$^{\phantom{***}}$
& \nominus{}0.151$^{*\phantom{**}}$
& \nominus{}0.043$^{\phantom{***}}$
& \niceminus{}0.024$^{\phantom{***}}$ \\

Qwen
& \niceminus{}0.148$^{*\phantom{**}}$
& \niceminus{}0.075$^{\phantom{***}}$
& \niceminus{}0.037$^{\phantom{***}}$
& \nominus{}0.056$^{\phantom{***}}$
& \nominus{}0.001$^{\phantom{***}}$
& \nominus{}0.146$^{*\phantom{**}}$
& \niceminus{}0.168$^{**\phantom{*}}$ \\

Swallow
& \niceminus{}0.033$^{\phantom{***}}$
& \niceminus{}0.295$^{***}$
& \niceminus{}0.076$^{\phantom{***}}$
& \niceminus{}0.149$^{*\phantom{**}}$
& \niceminus{}0.137$^{*\phantom{**}}$
& \nominus{}0.399$^{***}$
& \niceminus{}0.215$^{***}$ \\
\bottomrule
\end{tabular}
\caption{Pearson's correlation coefficients of each PoS, log frequency (BCCWJ) and polysemy (having more than one reference sense) with each model's zero-shot overall score. Adnominals (a single-word PoS) are excluded from the analysis. Significance was assessed using permutation tests with 10,000 permutations. Reported $p$-values are uncorrected for multiple comparisons: $^{*} p<0.05$, $^{**} p<0.01$, $^{***} p<0.001$.}
\label{tab:pcc-overall}
\end{table*}

\begin{table*}[htbp]
  \centering
  \small
  \begin{tabular}{lcccc}
    \toprule
    PoS & Words & Polysemous & log$_{10}$ Freq.\\
    \midrule
    
    Adnominal
        & \phantom{00}1   & 100\% & \niceminus{}3.4\phantom{\textpm{}0.0} \\
    Adverb
        & \phantom{0}12  & \phantom{0}50\%  & \niceminus{}4.2{\textpm{}0.8} \\
    \textit{i}-adjective
        & \phantom{0}21  & \phantom{0}62\%  & \niceminus{}4.5{\textpm{}0.6} \\
    \textit{na}-adjective
        & \phantom{0}14  & \phantom{00}7\%   & \niceminus{}4.7{\textpm{}0.5} \\
    Noun
        & 116 & \phantom{0}28\%  & \niceminus{}4.5{\textpm{}0.7} \\
    Verb
        & \phantom{0}86  & \phantom{0}58\%  & \niceminus{}4.4{\textpm{}0.7} \\
    \midrule
    All
        & 250 & \phantom{0}41\%  & \niceminus{}4.5{\textpm{}0.7} \\
        
    \bottomrule
  \end{tabular}
  \caption{Ratio of polysemous words (with more than one reference sense) and logarithm of frequency (mean and standard deviation of the logarithm of relative frequency in BCCWJ) by PoS in the test set of D3J.}
  \label{tab:polysemy-freq-by-pos}
\end{table*}

\begin{table*}[htbp]
  \centering\small
\begin{tabular}{llcccc}
\toprule
Model & Criterion & $R^2$ & $\Delta R^2_{\mathrm{freq}}$ & $\Delta R^2_{\mathrm{poly}}$ & $\Delta R^2_{\mathrm{PoS}}$ \\
\midrule
Claude Sonnet 4.6 & Overall & \underline{0.165} & 0.052$^{***}_{-}$ & 0.016$^{*\phantom{**}}_{-}$ & 0.071$^{***}$ \\
 & Precision & 0.063 & 0.029$^{**\phantom{*}}_{-}$ & 0.030$^{**\phantom{*}}_{+}$ & 0.019$^{\phantom{***}}$ \\
& Recall & \underline{0.175} & 0.000$^{\phantom{***}}$ & \underline{0.128}$^{***}_{-}$ & 0.009$^{\phantom{***}}$ \\
 & Specificity & 0.026 & 0.003$^{\phantom{***}}$ & 0.009$^{\phantom{***}}$ & 0.013$^{\phantom{***}}$ \\
 & Compliance & \underline{0.146} & 0.039$^{***}_{-}$ & 0.009$^{\phantom{***}}$ & 0.085$^{***}$ \\
\midrule
GPT-5.5 & Overall & \underline{0.121} & 0.080$^{***}_{-}$ & 0.015$^{*\phantom{**}}_{+}$ & 0.038$^{*\phantom{**}}$ \\
 & Precision & 0.045 & 0.012$^{\phantom{***}}$ & 0.013$^{\phantom{***}}$ & 0.026$^{\phantom{***}}$ \\
 & Recall & 0.075 & 0.028$^{**\phantom{*}}_{-}$ & 0.011$^{\phantom{***}}$ & 0.011$^{\phantom{***}}$ \\
 & Specificity & 0.066 & 0.028$^{**\phantom{*}}_{-}$ & 0.022$^{*\phantom{**}}_{+}$ & 0.030$^{\phantom{***}}$ \\
 & Compliance & 0.075 & 0.040$^{**\phantom{*}}_{-}$ & 0.000$^{\phantom{***}}$ & 0.021$^{\phantom{***}}$ \\
\midrule
Qwen & Overall & 0.051 & 0.016$^{*\phantom{**}}_{-}$ & 0.000$^{\phantom{***}}$ & 0.029$^{\phantom{***}}$ \\
 & Precision & 0.054 & 0.005$^{\phantom{***}}$ & 0.002$^{\phantom{***}}$ & 0.048$^{*\phantom{**}}$ \\
 & Recall & \underline{0.264} & 0.000$^{\phantom{***}}$ & \underline{0.207}$^{***}_{-}$ & 0.004$^{\phantom{***}}$ \\
 & Specificity & 0.043 & 0.021$^{*\phantom{**}}_{-}$ & 0.015$^{*\phantom{**}}_{+}$ & 0.011$^{\phantom{***}}$ \\
 & Compliance & 0.089 & 0.006$^{\phantom{***}}$ & 0.046$^{***}_{+}$ & 0.051$^{*\phantom{**}}$ \\
\midrule
Swallow & Overall & \underline{0.209} & 0.003$^{\phantom{***}}$ & 0.048$^{***}_{-}$ & \underline{0.123}$^{***}$ \\
 & Precision & 0.063 & 0.011$^{\phantom{***}}$ & 0.006$^{\phantom{***}}$ & 0.047$^{*\phantom{**}}$ \\
 & Recall & \underline{0.315} & 0.009$^{\phantom{***}}$ & \underline{0.224}$^{***}_{-}$ & 0.018$^{\phantom{***}}$ \\
 & Specificity & 0.053 & 0.017$^{*\phantom{**}}_{-}$ & 0.007$^{\phantom{***}}$ & 0.023$^{\phantom{***}}$ \\
 & Compliance & \underline{0.206} & 0.000$^{\phantom{***}}$ & 0.007$^{\phantom{***}}$ & \underline{0.184}$^{***}$ \\
\bottomrule
\end{tabular}
  \caption{Unique contributions of log frequency (BCCWJ), polysemy (having more than one reference sense), and PoS
    to zero-shot performance by model and criterion ($n = 249$, after exclusion of adnominals, a single-word PoS). For significance, see \cref{tab:pcc-overall}. Frequency and polysemy contribution subscripts show the coefficient sign for significant effects. Values larger than 0.1 are underlined.}
  \label{tab:regression-analysis}
\end{table*}

\begin{table*}[htbp]
  \centering\small
\begin{tabular}{llccc}
\toprule
 & & Polysemous & Senses & Max. Senses \\
\midrule
Reference  & & 41\% & 1.7{\textpm{}1.0} & \phantom{0}7 \\
\midrule
Zero-shot & Claude Sonnet 4.6 & 60\% & 2.0{\textpm{}1.1} & \phantom{0}7 \\
 & GPT-5.5 & 78\% & 2.6{\textpm{}1.5} & \phantom{0}8 \\
 & Qwen & 83\% & 2.6{\textpm{}1.2} & 10 \\
 & Swallow & 76\% & 1.8{\textpm{}0.6} & \phantom{0}4 \\
\midrule
Few-shot & Claude Sonnet 4.6 & 54\% & 1.8\textpm{}0.9 & \phantom{0}6 \\
 & GPT-5.5 & 73\% & 2.5\textpm{}1.4 & \phantom{0}9 \\
 & Qwen & 51\% & 1.6\textpm{}0.5 & \phantom{0}4 \\
 & Swallow & 46\% & 1.6\textpm{}0.6 & \phantom{0}5 \\
\midrule
Wiktionary &  & 72\% & 3.6{\textpm{}3.4} & 22 \\
\bottomrule
\end{tabular}
  \caption{Polysemy statistics: proportion of polysemous words, mean, standard deviation, and maximum number of senses. For few-shot prompting, we compute an average of three runs, where the standard deviation only reflects the dispersion over the dataset, not across the runs.}
  \label{tab:polysemy-stats}\end{table*}


\end{document}